\documentclass[a4paper,fleqn]{cas-dc}
\usepackage[numbers]{natbib}

\usepackage{microtype}
\usepackage{arydshln}
\usepackage{bm}

\usepackage{pifont}
\providecommand{\cmark}{\mbox{\ding{51}}}
\providecommand{\xmark}{\mbox{\ding{55}}}
\usepackage{amsthm}

\let\casfirstname\firstname
\renewcommand{\firstname}{\textcolor{black}{\casfirstname}}
\newcommand{\myparagraph}[1]{\vspace{0.1em}\noindent\textbf{#1}}

\hypersetup{colorlinks=true,breaklinks=true}
\renewcommand{\printorcid}{}
\def\tsc#1{\csdef{#1}{\textsc{\lowercase{#1}}\xspace}}
\tsc{WGM}
\tsc{QE}

\begin{document}
\let\WriteBookmarks\relax
\def\floatpagepagefraction{1}
\def\textpagefraction{.001}
\shorttitle{}  
\shortauthors{}  
\title [mode = title]{Text-Guided Modality Correction for Open-Vocabulary Object Detection}  
\author[1]{Yu Quan}
\ead{yuquan@xzhmu.edu.cn}

\author[2]{Dong Zhang}
\ead{dongzhang@njust.edu.cn}

\author[1]{Ping Gong}
\ead{gongping@xzhmu.edu.cn}

\author[3]{Peng Huang}
\ead{pengh0310@njfu.edu.cn}

\author[3]{JinHui Tang}
\ead{tangjh@njfu.edu.cn}
\cormark[2]


\affiliation[1]{organization={School of Medical Imaging, 
                              Xuzhou Medical University},
                city={Xuzhou},
                postcode={221004},
                country={China}}

\affiliation[2]{organization={School of Computer Science and Engineering,
                              Nanjing University of Science and Technology},
                city={Nanjing},
                postcode={210094},
                country={China}}

\affiliation[3]{organization={College of Information Science and Technology,
                              Nanjing Forestry University},
                city={Nanjing},
                postcode={210037},
                country={China}}
                
\cortext[2]{Corresponding author}

\begin{abstract}
Open-vocabulary object detection (OVD) is essential for detecting objects beyond the categories seen during training. Establishing vision–language associations is crucial for recognizing unseen categories, while integrating complementary visual features across levels supports detection at different scales. However, existing methods often provide insufficient textual guidance for object regions, rely on limited cross-modal alignment strategies, and underuse complementary features across levels, limiting their ability to detect unseen categories and objects of varying scales.
To address these limitations, we propose a Text-Guided Modality Correction Network (TMCNet) that incorporates textual guidance into region feature extraction, cross-modal alignment, and multi-scale decoding to strengthen region–category semantic correspondence and promote complementary feature fusion across levels. 
First, a text-guided modality encoder incorporates textual semantics into queries to guide regional feature aggregation, reducing semantic ambiguity in object representations. Second, a cross-modal alignment module strengthens the correspondence between region representations and category semantics through interactions between visual and textual features. Finally, a progressive text correction decoder combines a top-down pathway with lateral connections to progressively fuse text-guided deep semantic features with shallow visual features at the corresponding scales, preserving both semantic information and spatial detail for objects of different sizes. 
Experiments on two public datasets, COCO-OVD and LVIS-OVD, demonstrate that TMCNet achieves strong performance in open-vocabulary object detection using simple text queries as guidance. Extending TMCNet to open-vocabulary segmentation also yields performance gains, demonstrating its cross-task applicability and extensibility.
\end{abstract}
\maketitle
\section{Introduction}
\label{introduction}
Open-vocabulary object detection~\cite{gu2021open,zareian2021open} recognizes and localizes novel categories using textual descriptions, without requiring their detection annotations during training. Unlike conventional detectors trained on a fixed category set, it matches region representations with category semantics to extend recognition beyond base categories. Vision-language pretraining provides semantic priors for this purpose, but transferring image-level knowledge to localized object representations remains challenging.

\begin{figure}[pos=!tbp]
\centering
\includegraphics[width=.48 \textwidth]{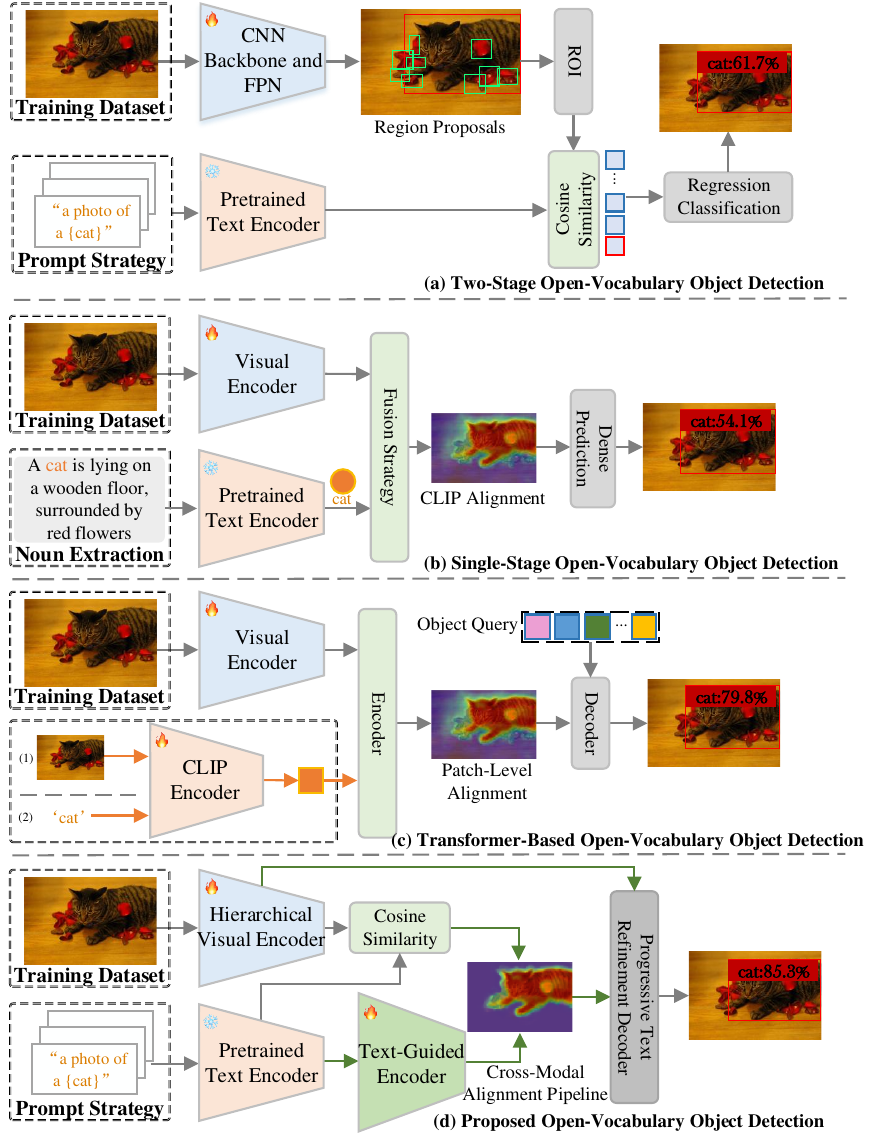}
\caption{Comparison of representative open-vocabulary object detection architectures. (a) Two-stage detectors extract proposal features for region--text classification and bounding-box regression. (b) Single-stage detectors fuse visual and textual features for dense classification and localization. (c) Transformer-based detectors use object queries to aggregate semantically aligned visual features for instance-level prediction. (d) TMCNet incorporates textual priors into query representations and combines cross-modal alignment with progressive multiscale decoding to guide visual feature construction.}
\label{fig1}
\end{figure}

Existing methods incorporate vision-language knowledge into several detection architectures. Two-stage methods (Fig.~\ref{fig1}(a)) classify region proposals through region-text matching and refine their bounding boxes. OVR-CNN~\cite{zareian2021open} learns visual-semantic representations from image-caption pairs, while ViLD~\cite{gu2021open} distills pretrained knowledge into region features. PromptDet~\cite{feng2022promptdet} adapts text representations through regional prompt learning, and BARON~\cite{wu2023aligning} exploits contextual relationships among regions for alignment.
Single-stage methods (Fig.~\ref{fig1}(b)) integrate semantic knowledge into dense prediction. Existing approaches use hierarchical knowledge distillation~\cite{ma2022open}, vision-language path aggregation with region-text contrastive learning~\cite{cheng2024yolo}, or lightweight alignment with adaptive decision learning~\cite{liu2024yolo} to combine category generalization with efficient detection. Query-based Transformer detectors (Fig.~\ref{fig1}(c)) instead aggregate visual features through object queries~\cite{carion2020end}. OV-DETR~\cite{zang2022open} conditions detection on text or exemplar image embeddings. 
Despite these advances, image-level semantic alignment does not directly provide the spatial representations required for foreground-background discrimination and multiscale localization. 
MMC-Det~\cite{xu2024exploring} introduces localization guidance through cross-modal masked language modeling; and CLIPSelf~\cite{wu2023clipself} refines region representations through self-distillation. These approaches motivate integrating textual guidance into visual feature processing beyond category matching.

We propose the Text-Guided Modality Correction Network (TMCNet) to integrate category semantics into query construction, cross-modal alignment, and multiscale decoding (Fig.~\ref{fig1}(d)). The text-guided modality encoder combines cross-attention and masked max pooling to construct queries with textual priors. The cross-modal alignment pipeline uses the aligned queries to correct lateral visual features and fuses them with projected vision-text association features to form joint representations. The progressive text correction decoder then aggregates the corrected lateral features and scale-adapted joint representations with the current decoder features. This design combines category semantics with spatial detail across feature levels for open-vocabulary classification and localization.

The main contributions of this paper are as follows:
\begin{itemize}
\item[$\bullet$] We propose a text-guided modality encoder that incorporates semantic priors into region feature extraction through interactions between text tokens and object queries, strengthening textual guidance for object representation learning.
\item[$\bullet$]  We design a cross-modal alignment module that combines modality projection with region-level vision–language interaction to strengthen region–category semantic correspondence and provide joint representations for subsequent decoding.
\item[$\bullet$] We develop a progressive text correction decoder that progressively fuses text-guided deep semantic features with multi-scale shallow visual features, combining semantic information and spatial detail across feature levels.
\item[$\bullet$] We evaluate detection performance on public benchmarks and extend the method to segmentation to assess its applicability across tasks.
\end{itemize} 
\vspace{-3mm}

\section{RELATED WORK}
\subsection{Open-Vocabulary Object Detection (OVD)}
\label{II-A}
Open-vocabulary object detection uses vision-language knowledge to detect novel categories without their bounding box annotations during training. OVR-CNN~\cite{zareian2021open} learns visual-semantic representations from image-caption pairs, while ViLD~\cite{gu2021open} transfers pretrained knowledge through region-level distillation. For query-based detection, OV-DETR~\cite{zang2022open} conditions predictions on text or exemplar image embeddings. 
Beyond semantic transfer, CAKE~\cite{ma2025cake} combines knowledge distillation with category-generalizable proposals to reduce novel-object suppression. 
WeDetect~\cite{fu2026wedetect} performs region-text retrieval with a fusion-free dual-tower architecture.
These methods address semantic transfer, novel-object proposal generation, and matching efficiency. TMCNet focuses on incorporating category semantics into visual feature construction and multiscale decoding. Its text-guided modality encoder constructs queries with textual priors, which the cross-modal alignment pipeline uses for lateral feature correction and joint feature modeling. The progressive text correction decoder integrates these outputs across scales to provide features for object classification and localization.

\subsection{Text-Guided Representation Learning and Cross-Modal Alignment}
\label{II-B}
Text-guided representation learning introduces category semantics into visual recognition. 
MMC-Det~\cite{xu2024exploring} models regional context through cross-modal masked language modeling to guide novel-object localization, while SRA-Det~\cite{yang2026sradet} extracts and matches multiple semantic facets from token-level text features for fine-grained recognition. These methods exploit complementary category information, regional context, and textual attributes. Building on these directions, TMCNet incorporates textual priors into query representations to guide subsequent visual feature processing. Its text-guided modality encoder combines cross-attention between object queries and text tokens with masked max pooling to construct semantic queries for feature correction and aggregation.

Cross-modal alignment further connects visual representations with category semantics. PromptDet~\cite{feng2022promptdet} adapts text representations through regional prompt learning, whereas BARON~\cite{wu2023aligning} exploits relationships among contextually related regions for vision-language alignment. CLIPSelf~\cite{wu2023clipself} aligns image-level and region-level visual representations through self-distillation. 
These methods improve alignment through text adaptation, contextual modeling, or visual representation refinement. TMCNet integrates alignment with multiscale feature decoding: its cross-modal alignment pipeline transforms TGME queries into semantic guidance for lateral feature correction and fuses them with projected vision-text association features. The resulting joint features support progressive aggregation in the decoder, extending textual guidance to multiple feature scales.

\subsection{Multi-Scale Feature Fusion}
\label{II-C}
Multi-scale feature fusion combines shallow spatial detail with deep semantic information for object recognition and localization. FPN~\cite{lin2017feature} propagates semantic information through a top-down pathway and lateral connections, while PANet~\cite{liu2018path} adds a bottom-up pathway to strengthen localization cues at higher levels. For open-vocabulary detection, YOLO-World~\cite{cheng2024yolo} integrates category semantics into hierarchical visual features through a re-parameterizable vision-language path aggregation network.
Open-vocabulary segmentation also benefits from combining semantic correspondence with spatial detail. CAT-Seg~\cite{cho2024catseg} aggregates vision-language similarity costs for dense prediction, whereas SED~\cite{xie2024sed} combines hierarchical encoding with progressive decoding. CLIPer~\cite{sun2025cliper} uses features and attention from early CLIP layers to improve spatial coherence and diffusion-model attention maps to refine local detail.
TMCNet incorporates textual guidance into both lateral feature correction and progressive aggregation. Its progressive text correction decoder combines the current decoder features with semantic-query-corrected lateral features and scale-adapted joint features from the cross-modal alignment pipeline. This design integrates category semantics with spatial information across feature levels to support object classification and localization.

\section{Methods}
\begin{figure*}[pos=t]
\centering
\includegraphics[width=.90\textwidth]{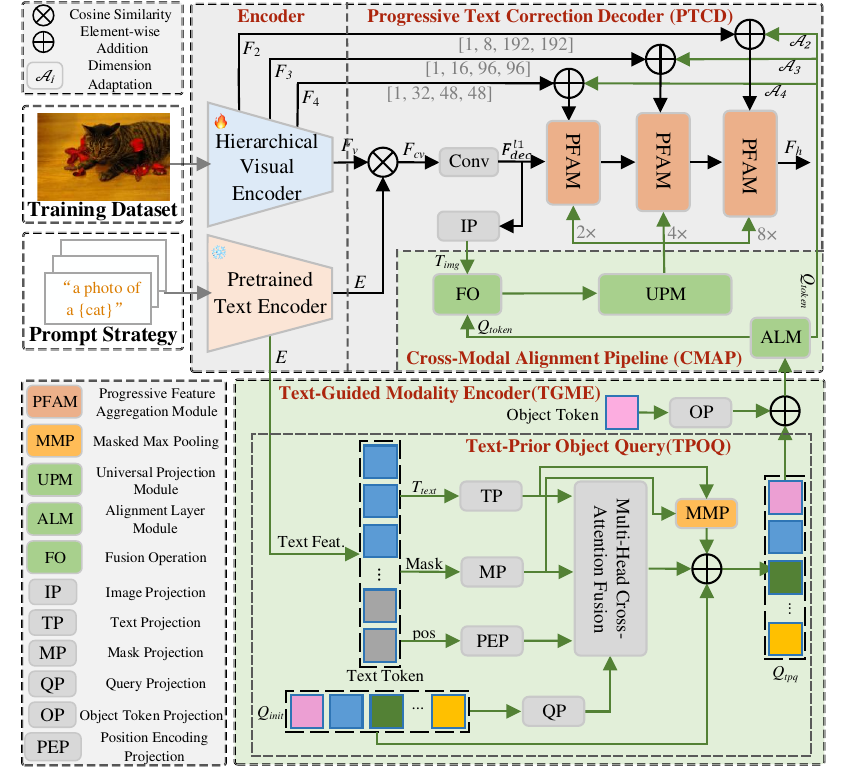}
\caption{Overview of TMCNet. A hierarchical visual encoder and a pretrained text encoder extract visual and category representations, respectively. The text-guided modality encoder (TGME) constructs queries with textual priors. The cross-modal alignment pipeline (CMAP) uses these queries to guide lateral feature correction and generate joint vision-language features. The progressive text correction decoder (PTCD) integrates both outputs through multiscale aggregation, producing features for object classification and localization.}
\label{fig2}
\end{figure*}

This section presents the architecture and components of the Text-Guided Modality Correction Network (TMCNet). Section~\ref{sec:3:1} provides an overview of the TMCNet and its processing pipeline. Sections~\ref{sec:3:2}–\ref{sec:3:4} then detail the text-guided modality encoder, cross-modal alignment pipeline, and progressive text correction decoder, respectively.  Finally, Section~\ref{sec:3:5} describes the extension of TMCNet to open-vocabulary semantic segmentation.

\subsection{Overall architecture}\label{sec:3:1}
Existing open-vocabulary object detection methods rely on pretrained text representations for category matching but underuse category semantics in region feature extraction and across visual hierarchies. Vision–text similarity alone also inadequately captures correspondences between local visual cues and category semantics. Inspired by recent advances in open-vocabulary detection and segmentation~\cite{xie2024sed,cheng2024yolo}, we propose the TMCNet, which integrates text information into object representations through semantically guided queries, region-level cross-modal interactions, and progressive feature fusion. As shown in Fig.~\ref{fig2}, TMCNet comprises a hierarchical visual encoder, a pretrained text encoder, a Text-Guided Modality Encoder (TGME), a Cross-Modal Alignment Pipeline (CMAP), a Progressive Text Correction Decoder (PTCD), and a detection head, jointly supporting visual–text encoding, semantic guidance, cross-modal fusion, and multi-scale prediction.

Specifically, in the visual branch, ConvNeXt~\cite{liu2022convnet} extracts four feature levels $\mathbf{F}_i$ ($i=2,3,4,5$) from an input image $\mathbf{I}\in\mathbb{R}^{H\times W\times3}$ at $1/4$, $1/8$, $1/16$, and $1/32$ resolution, respectively. Features $\mathbf{F}_2$, $\mathbf{F}_3$, and $\mathbf{F}_4$ supply multi-scale spatial information to the decoder via lateral connections. An MLP projects $\mathbf{F}_5$ into $\mathbf{F}_v\in\mathbb{R}^{H_v\times W_v\times D_t}$, matching the text embedding dimension for vision–text association, where $H_v=H/32$, $W_v=W/32$, and $D_t$ is the text embedding dimension.
In the text branch, prompt templates~\cite{cho2024catseg,liang2023open,gu2021open} generate $P$ descriptions per category, such as “a photo of a {class}.” For category $n$, these descriptions form $\mathcal{S}_n=\{s_{n,p}\}_{p=1}^{P}$. A pretrained text encoder produces embeddings $E = \{E_1,\dots, E_N\}\in R^{N\times P\times D_t}$, where $N$ is the number of categories. Cosine similarities between visual features at each spatial location and category text embeddings form a dense vision–text cost map $\mathbf{F}_{cv}$:

\begin{equation}
\label{eq1}
\begin{aligned}
\textbf{F}_{cv}(i,j,n,p) = \frac{{\textbf{F}_{v}(i,j)E(n,p)}}{\left \| {\textbf{F}}_v(i,j) \right \| \left \| E(n,p) \right \|}
\end{aligned},
\end{equation}
Here, $(i,j)$ denotes a spatial location, while $n$ and $p$ index categories and prompt templates, respectively. The cost map $\mathbf{F}_{cv}\in\mathbb{R}^{H_v\times W_v\times N\times P}$ encodes initial semantic associations between spatial locations and category descriptions. A convolutional embedding maps $\mathbf{F}_{cv}$ to the initial decoder features $\mathbf{F}_{dec}^{(1)}\in\mathbb{R}^{H_v\times W_v\times N\times D}$, where $D$ denotes the decoder feature dimension.

TGME incorporates category semantics into object queries to guide visual feature aggregation. The Text-Prior Object Query Module (TPOQ) uses multi-head cross-attention with padding masks, positional embeddings  and $\mathbf{Q}{init}$ to aggregate text features $\mathbf{T}{text}$ into learnable object queries. In parallel, masked max pooling extracts a text representation from valid tokens. The attention output, pooled representation, and initial queries are summed to obtain text-prior object queries $\mathbf{Q}{tpq}$, which are then combined with projected learnable object tokens to form $\mathbf{Q}{token}$ for subsequent cross-modal feature fusion. Details of TGME are provided in Section~\ref{sec:3:2}.
CMAP aligns and fuses semantically guided queries with vision-text association features to support decoding through two complementary branches. First, the Alignment Layer Module (ALM) transforms $\mathbf{Q}_{token}$ into a representation compatible with the visual branch, while the image projection layer maps $\mathbf{F}_{dec}^{l1}$ to $\mathbf{T}_{img}$. The aligned queries retain category semantics for lateral feature correction and are also combined with $\mathbf{T}_{img}$ through the Fusion Operation (FO). The fused representation then passes through the Universal Projection Module (UPM) to produce joint vision-language features for decoder feature aggregation. These two branches incorporate text semantics into lateral feature correction and feature aggregation, respectively. Details of CMAP are provided in Section~\ref{sec:3:3}.
Finally, PTCD progressively fuses $\mathbf{F}_4$, $\mathbf{F}_3$, and $\mathbf{F}_2$ through a top-down pathway with lateral connections. At each stage, the ALM representation is adapted to the spatial and channel dimensions of the corresponding visual features and added element-wise for semantic correction. The Progressive Feature Aggregation Module (PFAM)~\cite{xie2024sed} then fuses these corrected features with the current decoder features and UPM joint features to restore spatial resolution. This process combines high-level category semantics with fine spatial details across feature levels. The resulting features $\mathbf{F}_h$ are passed to a detection head for category prediction and bounding box regression, supervised by classification and regression losses. Details of PTCD are provided in Section~\ref{sec:3:4}.

\subsection{TGME: Text-Guided Modality Encoder}\label{sec:3:2}
The TGME incorporates category semantics into learnable object queries through the TPOQ, providing semantic priors for subsequent visual feature aggregation. As shown in Fig.~\ref{fig2}, the encoder first aggregates text information using multi-head cross-attention and masked max pooling to generate text-prior object queries. These queries are then fused with projected learnable object tokens to form semantically guided representations.

\myparagraph{Multi-modal Token Construction.}
The text encoder outputs embeddings $E=\{E_1,\dots,E_N\}\in\mathbb{R}^{N\times P\times D_t}$, which are mapped to text representations $L_{text}=\{l_1,\dots,l_N\}\in\mathbb{R}^{N\times P\times D_t}$. Here, $N$, $P$, and $D_t$ denote the number of categories, prompt templates per category, and text embedding dimension, respectively. The corresponding padding mask $L_{mask}=\{l_m^1,\dots,l_m^N\}\in\mathbb{R}^{N\times P\times M}$ is derived from input tokenization and padding to identify valid tokens, where $M$ is the maximum text length. The text tokens are then linearly projected along the channel dimension to obtain $T_{text}\in\mathbb{R}^{N_{text}\times C}$, with $N_{text}$ denoting the number of tokens used in subsequent interactions and $C$ the projected channel dimension.
A learnable object token $T_{obj}$ is introduced for the query token branch. Meanwhile, the image projection layer in CMAP maps $F_{dec}^{l1}$ to $T_{img}=\{t^1,\dots,t^N\}\in\mathbb{R}^{N\times C\times D}$ using parameters independent of the text projection.
The object, image, and text tokens are collectively denoted by $\textbf{Query}$, with their corresponding padding masks denoted by $\textbf{Mask}$:
\begin{equation}
\label{eq2}
\begin{aligned}
\textbf{Query} = \{T_{obj}, t^1, \dots, t^N, l^1, \dots, l^N \},\\
\textbf{Mask} = \{T_{objm}, t^1_m, \dots, t^N_m, l^1_m, \dots, l^N_m \}.\\
\end{aligned}
\end{equation}
where $T_{objm}$, $t_m^n$, and $l_m^n$ denote the padding masks corresponding to $T_{obj}$, $t^n$, and $l^n$, respectively.

\myparagraph{Text-Prior Object Queries.}
The TPOQ incorporates input category semantics into learnable object queries $Q_{init}$ to guide subsequent visual feature aggregation. As shown in Fig.~\ref{fig2}, it first projects the text features and adapts the corresponding padding mask:
\begin{equation}
\label{eq3}
\begin{aligned}
Z_{text} &= \operatorname{TP}(T_{text}),\\
M_{text} &= \operatorname{MP}(Mask).
\end{aligned}
\end{equation}
Here, $\operatorname{TP}$ projects the text features to $Z_{text}$, while $\operatorname{MP}$ adapts the text padding mask $Mask$ to obtain $M_{text}$, which identifies valid and padded positions in the text sequence.
Using these inputs, multi-head cross-attention aggregates query-specific text information, with projected object queries as queries, position-encoded text features as keys, and $Z_{text}$ as values. In parallel, masked max pooling extracts a global text representation from valid positions and broadcasts it across queries. Both outputs are added to $Q_{init}$ to obtain text-prior object queries:
\begin{equation}
\label{eq4}
\begin{aligned}
Q_{tpq}
={}&
\operatorname{MCA}\Big(
\operatorname{QP}(Q_{init}),
Z_{text}+\operatorname{PEP}(pos),
Z_{text};
M_{text}
\Big)\\
&+
\operatorname{Broadcast}_{N_q}
\Big(
\operatorname{MMP}(Z_{text},M_{text})
\Big)+Q_{init}.
\end{aligned}
\end{equation}
Here, $\operatorname{MCA}(Q,K,V;M)$ denotes multi-head cross-attention with queries $Q$, keys $K$, values $V$, and a mask $M$ that excludes padded positions. $\operatorname{QP}$ and $\operatorname{PEP}$ are independent linear projections of the queries and text positional encodings $pos$, respectively. $\operatorname{MMP}$ computes channel-wise maxima over valid text positions, and $\operatorname{Broadcast}_{N_q}$ expands the pooled representation to match the $N_q$ queries. The resulting $Q_{tpq}$ combines query-specific text information with shared global semantics for subsequent cross-modal interactions.

\myparagraph{Query Token Branch.}
This branch combines the learnable object token $T_{obj}\in R^{1\times C}$ with the text-prior queries $Q_{tpq}$ generated by TPOQ. Specifically, $OP$ linearly projects $T_{obj}$ into the query feature space, and the projected token is broadcast across queries and added element-wise to $Q_{tpq}$. An MLP then transforms the fused features into semantically guided representations:
\begin{equation}
\label{eq5}
\begin{aligned}
\textbf{Q}_{token} = \textbf{MLP}(\textbf{OP}(T_{obj}) +  \textbf{TPQ}(\textbf{TP}(T_{text}), Q_{obj}))
\end{aligned}.
\end{equation}
Here, $TPQ$ denotes TPOQ, and $Q_{obj}$ refers to the initial queries $Q_{init}\in R^{N_{oq}\times C}$, where $N_{oq}$ is the number of object queries. The resulting $Q_{token}$ carries text-derived category semantics and is passed to CMAP to guide visual feature fusion.

\subsection{CMAP: Cross-Modal Alignment Pipeline}\label{sec:3:3}
TGME queries encode category semantics, while the initial decoder features $F_{dec}^{l1}$ capture associations between spatial locations and category descriptions. CMAP brings these representations together to guide visual feature aggregation. As shown in Fig.~\ref{fig2}, it comprises an Alignment Layer Module (ALM), a Fusion Operation (FO), and a Universal Projection Module (UPM). The ALM output supplies category semantics for lateral feature correction and also enters FO and UPM to form joint features for decoder aggregation.

\myparagraph{Alignment Layer Module.}
To enable interaction between the two representations, ALM uses a two-layer MLP to map the TGME queries $Q_{token}\in\mathbb{R}^{N_q\times C}$ into a shared feature space:
\begin{equation}
\label{eq6}
\bar{Q}_{token}
=
\operatorname{ALM}(Q_{token})
\in\mathbb{R}^{N_q\times C}.
\end{equation}
Here, $N_q$ is the number of queries, $C$ is the shared channel dimension, and $\bar{Q}_{token}$ denotes the transformed queries. This mapping prepares the queries for cross-modal interaction; their correspondence with visual content is learned through subsequent attention and detection supervision.

\myparagraph{Fusion Operation.}
FO uses the aligned queries to aggregate vision-text association features. First, the spatial and category dimensions of $F_{dec}^{l1}\in\mathbb{R}^{H_v\times W_v\times N\times D}$ are flattened, and the image projection layer $IP$ maps the resulting features to $C$ channels:
\begin{equation}
\label{ch5:eq6}
T_m
=
\operatorname{IP}
\left(
\operatorname{Flatten}(F_{dec}^{l1})
\right)
\in\mathbb{R}^{L_v\times C},
\end{equation}
where $H_v$ and $W_v$ are the spatial dimensions, $N$ is the number of categories, and $D$ is the decoder feature dimension. The resulting $L_v=H_vW_vN$ tokens each represent a spatial location-category pair. Their original indices are retained for subsequent layout restoration.
The aligned queries and modality tokens are then concatenated into a joint sequence:
\begin{equation}
\label{ch5:eq7}
T_{cat}
=
\operatorname{Concat}_{seq}
\left(
\bar{Q}_{token},T_m
\right)
\in\mathbb{R}^{(N_q+L_v)\times C}.
\end{equation}
Here, $\operatorname{Concat}_{seq}$ denotes sequence-wise concatenation, placing the $N_q$ query tokens before the $L_v$ modality tokens. To enrich the queries with relevant visual context, scaled dot-product attention uses $\bar{Q}_{token}$ as queries and $T_m$ as keys and values:
\begin{equation}
\label{ch5:eq8}
A
=
\operatorname{Softmax}
\left(
\frac{\bar{Q}_{token}T_m^{\top}}{\sqrt{C}}
\right)
\in\mathbb{R}^{N_q\times L_v}.
\end{equation}
Softmax normalizes scores over the $L_v$ modality tokens, yielding query-specific weights for aggregation:
\begin{equation}
\label{ch5:eq9}
T_{C\text{-}Attention}
=
AT_m
\in\mathbb{R}^{N_q\times C}.
\end{equation}
The resulting context features update only the query positions through residual addition, leaving the modality tokens unchanged:
\begin{equation}
\label{ch5:eq10}
\begin{aligned}
\tilde{T}_m
&=
T_{cat}
+
\operatorname{Concat}_{seq}
\left(
T_{C\text{-}Attention},
0_{L_v\times C}
\right)\\
&=
\operatorname{Concat}_{seq}
\left(
\bar{Q}_{token}+T_{C\text{-}Attention},
T_m
\right).
\end{aligned}
\end{equation}
where $0_{L_v\times C}$ is a zero tensor indicating that the residual update applies only to the query positions in the joint sequence.  $\tilde{T}_m$ retains the original vision--text association tokens alongside queries enriched with visual context.

\myparagraph{Universal Projection Module.}
UPM applies Transformer encoder layers to the fused sequence, modeling token interactions through self-attention and token-wise transformations while preserving sequence length:
\begin{equation}
\label{ch5:eq11}
\hat{T}_m
=
\operatorname{UPM}(\tilde{T}_m)
\in\mathbb{R}^{(N_q+L_v)\times C}.
\end{equation}
Joint self-attention allows modality tokens to incorporate query semantics and contextual information from other tokens. The last $L_v$ output tokens are then reshaped according to their original indices to recover $F_{joint}\in\mathbb{R}^{H_v\times W_v\times N\times C}$. After spatial and channel adaptation, these joint features are supplied to the Progressive Feature Aggregation Module at each decoding stage.

Together, the two CMAP branches provide complementary decoder inputs: $\bar{Q}_{token}$ modulates the lateral features $F_2$, $F_3$, and $F_4$ through dimensional adaptation, while $F_{joint}$ supplies cross-modal context for progressive aggregation. The decoder combines these inputs with its current features to restore spatial resolution, integrating high-level semantics with fine spatial details for category prediction and bounding box regression.

\subsection{PTCD: Progressive Text Correction Decoder}\label{sec:3:4}
As shown in Fig.~\ref{fig2}, PTCD progressively fuses hierarchical visual features through a top-down pathway with lateral connections. The two CMAP branches provide complementary guidance: ALM queries supply category semantics for lateral feature correction, while UPM joint features provide cross-modal context for progressive aggregation.

\myparagraph{Semantically Guided Lateral Feature Correction.}
Since ALM queries have a sequence structure, dimensional adaptation first maps them to representations matching the spatial layout and channel dimension of each lateral feature. These representations are then added element-wise to $F_4$, $F_3$, and $F_2$, respectively, incorporating category semantics before decoder aggregation. This mapping converts query representations into spatially compatible features rather than directly upsampling the query sequence.

\myparagraph{Progressive Aggregation Guided by Joint Cross-Modal Features.}
The UPM branch complements lateral correction with joint features that encode spatial location--category associations. After their spatial layout is restored, these features are resampled and channel-projected to match each decoding stage, providing PFAM~\cite{xie2024sed} with semantic and visual context at multiple scales.

\myparagraph{Multi-Scale Feature Fusion and Prediction.}
Starting from $F_{dec}^{l1}$, PTCD performs three aggregation stages corresponding to $F_4$, $F_3$, and $F_2$. At each stage, the current decoder features are resized to match the lateral resolution. PFAM~\cite{xie2024sed} then fuses them with the semantically corrected lateral features and the adapted UPM features. The output proceeds to the next stage to incorporate higher-resolution information.
This process propagates high-level semantics along the top-down pathway while adding spatial details through lateral connections. The final features $F_h$ are passed to the detection head for category prediction and bounding box regression.

\subsection{Extension to Open-Vocabulary Semantic Segmentation}\label{sec:3:5}
Open-vocabulary semantic segmentation~\cite{li2023llava,liu2023visual,liu2023medical} faces similar challenges: limited adaptability of pretrained text representations, underuse of textual semantics, and a mismatch between global image-text alignment and pixel-level prediction~\cite{xie2024sed}. To address these challenges, we integrate TMCNet into the SED~\cite{xie2024sed} and OVSeg~\cite{liang2023open} frameworks to enhance pixel-level representations through text-guided feature correction, cross-modal alignment, and multi-scale fusion. Experiments show improved segmentation accuracy in both frameworks, supporting the effectiveness of TMCNet for dense prediction and its applicability across tasks.

\section{Experiments}
We evaluate TMCNet on the COCO-OVD~\cite{zareian2021open} and LVIS-OVD~\cite{gu2021open} datasets and examine its extension to open-vocabulary semantic segmentation. Sections~\ref{sec4:1} and~\ref{sec4:2} describe the dataset, evaluation metrics, and implementation details. Section~\ref{sec4:3} evaluates TGME, CMAP, and PTCD through ablation studies and comparisons with alternative designs. Section~\ref{sec4:4} presents quantitative comparisons with recent state-of-the-art detection methods and qualitative visualizations. Finally, Section~\ref{sec4:5} evaluates the segmentation extension using SED~\cite{xie2024sed} and OVSeg~\cite{liang2023open} as baselines.

\subsection{Datasets and Evaluation Metrics}\label{sec4:1}
\myparagraph{Datasets.}
We evaluate TMCNet on open-vocabulary object detection and semantic segmentation using the following datasets and protocols.
\begin{itemize}
\item Open-vocabulary object detection (OVD): Following prior work~\cite{zareian2021open,ma2025cake,gu2021open,lee2026hierarchical}, we use COCO-OVD and LVIS-OVD, built on MS-COCO~\cite{lin2014microsoft} and LVIS v$1.0$~\cite{gupta2019lvis}, respectively. We adopt the category splits and evaluation protocols of OVR-CNN~\cite{zareian2021open} for COCO-OVD and ViLD~\cite{gu2021open} for LVIS-OVD.
\item Open-vocabulary semantic segmentation (OVS): Following prior methods~\cite{cho2024catseg,xu2023side,xie2024sed,chen2026errseg}, we train on COCO-Stuff~\cite{caesar2018coco} and evaluate on ADE20K~\cite{zhou2019semantic}, PASCAL VOC~\cite{everingham2010pascal}, and PASCAL-Context~\cite{mottaghi2014role}. Evaluation covers the $150$-class and $847$-class settings of ADE20K (A-$150$ and A-$847$), the $20$-class setting of PASCAL VOC (PAS-$20$), and the $59$-class and $459$-class settings of PASCAL-Context (PC-$59$ and PC-$459$).
\end{itemize}

\myparagraph{Evaluation Metrics.}
We assess prediction accuracy using task-specific metrics and report computational efficiency for detection.
\begin{itemize}
\item Open-vocabulary object detection (OVD): On COCO-OVD, the primary metric is $\mathrm{AP}_{50}^{\mathrm{novel}}$, the average precision for novel categories at an IoU threshold of $0.5$. We also report $\mathrm{AP}_{50}^{\mathrm{base}}$ and $\mathrm{AP}_{50}^{\mathrm{all}}$ for base categories and the union of base and novel categories, respectively. On LVIS-OVD, $\mathrm{AP}_{r}$ for rare categories is the primary metric for novel-category detection, supplemented by $\mathrm{AP}_{c}$ for common categories, $\mathrm{AP}_{f}$ for frequent categories, and $\mathrm{mAP}$ across all categories. All LVIS-OVD AP metrics are averaged over IoU thresholds from $0.50$ to $0.95$ in increments of $0.05$.
\item Open-vocabulary semantic segmentation (OVS): We report mean Intersection over Union (mIoU) on all evaluation datasets, computed by averaging the IoU between predicted and ground-truth regions across evaluated categories.
\end{itemize}
For detection, we additionally report inference time (Inf.time), floating-point operations (FLOPs), and parameter count (Params.) to measure inference latency, computational cost, and model size, respectively.

\subsection{Implementation Details}\label{sec4:2}
\myparagraph{Baselines.}\label{sec4:2:1}
We build TMCNet on SED~\cite{xie2024sed} and OVSeg~\cite{liang2023open}, two open-vocabulary semantic segmentation frameworks. For detection, we adapt them into SED-OVD and OVSeg-OVD using the CLIP\_Trans head for instance classification and bounding box prediction, as described in Section~\ref{sec4:2:3}. For segmentation, we retain their original prediction mechanisms. Unless required for module integration or task adaptation, other settings follow the corresponding baseline.
\begin{itemize}
\item OVSeg-OVD: OVSeg~\cite{liang2023open} combines a MaskFormer-based segmentation model~\cite{cheng2021per} with mask-adapted CLIP and ensembles their class predictions for masked regions. We use ResNet-101c as the segmentation backbone and CLIP with a ViT-B/16~\cite{dosovitskiy2020image} visual encoder. Mask prediction and class ensembling are retained for segmentation, while the detection variant uses the task-specific head.
\item SED-OVD: SED~\cite{xie2024sed} combines a hierarchical visual encoder with a progressive fusion decoder to aggregate image-text cost features for pixel-level prediction. We use pretrained CLIP with a ConvNeXt-B~\cite{xie2024sed} visual encoder and its corresponding text encoder. The text embedding dimension is $D_t=640$, with $P=80$ prompt templates per category. The initial decoder features $F_{dec}^{l1}$ have $128$ channels.
\end{itemize}

\myparagraph{Visual Backbones and Encoders.}\label{sec4:2:2}
The visual encoders provide spatial and semantic representations according to the underlying framework:
\begin{itemize}
\item ResNet-101c: In the OVSeg-OVD framework, convolutional and residual stages extract spatial features for subsequent fusion and task-specific prediction.
\item ViT-B/16~\cite{dosovitskiy2020image}: The CLIP visual encoder embeds non-overlapping $16\times16$ patches and processes them with a classification token and positional embeddings through Transformer layers. The classification token aggregates image-level information, while patch tokens retain spatial correspondence. Unlike ResNet-101c, standard ViT-B/16 does not directly produce a hierarchical feature pyramid.
\item ConvNeXt-B~\cite{liu2022convnet}: In the SED-OVD framework, four convolutional stages provide hierarchical features for image-text matching and progressive fusion. For an input of size $H\times W$, with both dimensions divisible by $32$, the output strides are $4$, $8$, $16$, and $32$, with channel dimensions of $128$, $256$, $512$, and $1024$, respectively. In height-width-channel order, the feature shapes are $(H/4,W/4,128)$, $(H/8,W/8,256)$, $(H/16,W/16,512)$, and $(H/32,W/32,1024)$. These backbone channel dimensions differ from the projected image-text embedding dimension $D_t=640$.
\end{itemize}

\myparagraph{Detection Head and Task Adaptation.}\label{sec4:2:3}
The query-based CLIP\_Trans head uses learnable instance queries enriched with TGME semantics to attend to channel-projected PTCD features with positional embeddings. Through query self-attention and cross-attention to these visual features, a Transformer produces instance representations for classification and localization. The classification branch projects them into the category text embedding space and computes cosine similarities with category embeddings, while a class-agnostic MLP predicts normalized bounding box coordinates. Training uses bounding boxes and class labels from base categories. At inference, the head predicts class labels, confidence scores, and bounding boxes over a vocabulary containing both base and novel categories.
For segmentation, the proposed text-guided feature correction modules are integrated into SED and OVSeg while retaining their segmentation prediction mechanisms. The detection and segmentation models are trained with bounding box annotations and pixel-level semantic annotations, respectively.

\myparagraph{Training Settings.}\label{sec4:2:4}
Experiments run on two NVIDIA GeForce RTX $4090$ GPUs, each with $24$\,GB of memory. The OVSeg-OVD and SED-OVD detectors model use the following configurations.
\begin{itemize}
\item OVSeg-OVD: We follow the OVSeg training setup with an adjusted batch size, using random crops of $640\times640$ pixels, a global batch size of 8, and $120\,000$ iterations. AdamW~\cite{loshchilov2017decoupled} uses an initial learning rate of $6\times10^{-5}$ and weight decay of $1\times10^{-2}$, with polynomial learning rate decay.
\item SED-OVD: Distributed data parallel training uses random crops of $768\times768$ pixels and a global batch size of $2$, with one image per GPU and no gradient accumulation. AdamW~\cite{loshchilov2017decoupled} uses a base learning rate of $2\times10^{-4}$ and default weight decay of $1\times10^{-4}$; normalization and embedding layers are exempt from weight decay. We apply cosine learning rate decay without warm-up and clip the global gradient norm to $0.01$. The maximum training budget is $1\,200\,000$ iterations, with checkpoints saved every $5\,000$ iterations and evaluation every $10\,000$ iterations.
\end{itemize}

\myparagraph{Inference Settings.}\label{sec4:2:5}
Inference preprocessing follows the configuration of each model. For SED-OVD, images are resized with a target short-side length of 640 pixels and a maximum long-side length of $2560$ pixels while preserving the aspect ratio. Neither test-time augmentation nor sliding-window inference is used. Inference time (Inf.~time) is measured on a single NVIDIA GeForce RTX $4090$ GPU with $24$\,GB of memory, using FP16 precision and a batch size of $1$.

\begin{table*}[!thbp]
\caption{Ablation study of TGME, CMAP, and PTCD on two detection-adapted baselines on COCO-OVD. All results are reported in \%. $^{*}$ indicates the detection-adapted variants of OVSeg and SED. $\cmark$ and $\xmark$ indicate the presence and absence of the corresponding module, respectively. Bold indicates the best result for each baseline.}\label{Tab1}
\centering
\renewcommand\arraystretch{1.0}
\resizebox{0.85\linewidth}{!}{
\normalsize
\begin{tabular}{c|c|c|ccc|ccc}
\hline
\hline
\multirow{2}{*}{Detection-adapted Baseline} & \multirow{2}{*}{VLM} & \multirow{2}{*}{Feature backbone} & \multirow{2}{*}{TGME} & \multirow{2}{*}{CMAP} & \multirow{2}{*}{PTCD} & \multicolumn{3}{c}{COCO-OVD(\%)} \\
& & & & & & AP$_{50}^\textrm{novel}$ & AP$_{50}^\textrm{base}$ &  AP$_{50}^\textrm{all}$\\
\hline
\multirow{4}{*}{OVSeg-OVD$^{*}$} & ViT-B/16 & ResNet-101c	& \xmark & \xmark & \xmark & 33.70 & 53.70 & 49.30 \\
\multirow{4}{*} & ViT-B/16 & ResNet-101c & \cmark & \xmark & \xmark & 37.90$_{\color{red}\textrm{+4.20}}$ & 58.00	&52.80 \\
\multirow{4}{*} & ViT-B/16 & ResNet-101c & \xmark & \cmark & \xmark  &	35.10$_{\color{red}\textrm{+1.40}}$ & 55.20 & 51.40 \\
\multirow{4}{*} & ViT-B/16 & ResNet-101c & \cmark & \cmark & \xmark  &	38.50$_{\color{red}\textrm{+4.80}}$ & 58.20 & 53.00 \\
\multirow{4}{*} & ViT-B/16 & ResNet-101c & \cmark & \cmark & \cmark & \textbf{39.20}$_{\color{red}\textrm{+5.50}}$ & \textbf{58.50} & \textbf{53.30} \\
\cdashline{1-9}
\multirow{4}{*}{SED-OVD$^{*}$} & ConvNeXt-B & -	& \xmark & \xmark & \xmark & 36.10 & 54.20 & 49.90 \\
\multirow{4}{*} & ConvNeXt-B & - & \cmark & \xmark & \xmark & 40.40$_{\color{red}\textrm{+4.30}}$ & 58.50 & 54.10\\
\multirow{4}{*} & ConvNeXt-B & - & \xmark & \cmark & \xmark  & 37.70$_{\color{red}\textrm{+1.60}}$ &55.90 &	51.40 \\
\multirow{4}{*} & ConvNeXt-B & - & \cmark & \cmark & \xmark  &	41.00$_{\color{red}\textrm{+4.90}}$ & 59.50 & 54.70 \\
\multirow{4}{*} & ConvNeXt-B & - & \cmark & \cmark & \cmark & \textbf{41.60}$_{\color{red}\textrm{+5.50}}$ & \textbf{60.90}	& \textbf{55.40} \\
\hline\hline
\end{tabular}
}
\end{table*}
\begin{table*}[b]
\caption{Ablation results for different module configurations on LVIS-OVD. All results are reported in \%. $^{*}$ indicates the detection-adapted variants of OVSeg and SED. \cmark and \xmark indicate the presence and absence of the corresponding module, respectively. Bold indicates the best result for each baseline.}\label{Tab2}
\centering
\renewcommand\arraystretch{1.0}
\resizebox{0.85\linewidth}{!}{
\normalsize
\begin{tabular}{c|c|c|ccc|cccc}
\hline
\hline
\multirow{2}{*}{Detection-adapted Baseline} & \multirow{2}{*}{VLM} & \multirow{2}{*}{Feature backbone} & \multirow{2}{*}{TGME} & \multirow{2}{*}{CMAP} & \multirow{2}{*}{PTCD} & \multicolumn{4}{c}{LVIS-OVD(\%)} \\
& & & & & & AP$_r$ & AP$_c$ &  AP$_f$ & mAP \\
\hline
\multirow{4}{*}{OVSeg-OVD$^{*}$} & ViT-B/16 & ResNet-101c & \xmark & \xmark & \xmark & 22.40 & 28.30	& 36.00	& 32.40\\
\multirow{4}{*} & ViT-B/16 & ResNet-101c & \cmark & \xmark & \xmark & 24.90$_{\color{red}\textrm{+2.50}}$	& 30.70	& 38.20 & 34.50\\
\multirow{4}{*} & ViT-B/16 & ResNet-101c & \xmark & \cmark & \xmark  & 23.80$_{\color{red}\textrm{+1.40}}$ & 	29.50 & 37.40	& 33.70\\
\multirow{4}{*} & ViT-B/16 & ResNet-101c & \cmark & \cmark & \xmark  & 26.1$_{\color{red}\textrm{+3.70}}$ & 31.90 & 39.50	& 36.00\\
\multirow{4}{*} & ViT-B/16 & ResNet-101c & \cmark & \cmark & \cmark & \textbf{26.70}$_{\color{red}\textrm{+4.30}}$	& \textbf{37.10}	& \textbf{40.00}	& \textbf{36.50}\\
\cdashline{1-10}
\multirow{4}{*}{SED-OVD$^{*}$} & ConvNeXt-B & - & \xmark & \xmark & \xmark & 24.80 & 34.50	& 38.00	& 34.00\\
\multirow{4}{*} & ConvNeXt-B & - & \cmark & \xmark & \xmark & 25.90$_{\color{red}\textrm{+1.10}}$	& 35.80	& 39.20	& 35.30\\
\multirow{4}{*} & ConvNeXt-B & - & \xmark & \cmark & \xmark  &	25.30$_{\color{red}\textrm{+0.50}}$ & 34.90	& 38.60	& 34.80\\
\multirow{4}{*} & ConvNeXt-B & - & \cmark & \cmark & \xmark  &	27.40$_{\color{red}\textrm{+2.60}}$ & 36.70	& 40.40	& 36.60\\
\multirow{4}{*} & ConvNeXt-B & - & \cmark & \cmark & \cmark & \textbf{28.60}$_{\color{red}\textrm{+3.80}}$	& \textbf{38.10}	& \textbf{41.00} & \textbf{37.70}\\
\hline\hline
\end{tabular}}
\end{table*}
\subsection{Ablation Study}\label{sec4:3}
We conduct ablation studies on COCO-OVD and LVIS-OVD using detection-adapted OVSeg-OVD and SED-OVD as baselines. Tables~\ref{Tab1} and~\ref{Tab2} evaluate module configurations on COCO-OVD and LVIS-OVD, respectively. Table~\ref{Tab3} compares TGME with direct cosine similarity matching, feature fusion, and encoder-based interaction. With TGME enabled, Table~\ref{Tab4} examines alternative cross-modal alignment strategies, complemented by the visualizations in Fig.~\ref{fig4}. Finally, Table~\ref{Tab5} compares detection heads with TGME and CMAP enabled, including the CLIP+Transformer head.

\subsubsection{Effectiveness Across Baselines}
For each baseline, we evaluate TGME and CMAP alone, their combination, and the complete model with PTCD. This comparison assesses the individual contributions of TGME and CMAP and the additional benefit of PTCD when both modules are present.

\textbf{COCO-OVD.}
Table~\ref{Tab1} shows that TGME alone improves $\mathrm{AP}_{50}^{\mathrm{novel}}$ by $4.20$ and $4.30$ percentage points on the OVSeg-OVD and SED-OVD baselines, respectively, exceeding the gains of $1.40$ and $1.60$ points from CMAP alone. Combining TGME and CMAP raises novel-category AP to $38.50$\% and $41.00$\%. Adding PTCD further improves these scores by $0.70$ and $0.60$ points, yielding $39.20$\% and $41.60$\%, respectively. The complete model also achieves the highest base- and all-category AP, with $\mathrm{AP}_{50}^{\mathrm{all}}$ improving over the baselines by $4.00$ and $5.50$ points.

\textbf{LVIS-OVD.}
Table~\ref{Tab2} shows consistent improvements across both baselines. TGME alone increases $\mathrm{AP}_{r}$ by $2.50$ and $1.10$ percentage points for OVSeg-OVD and SED-OVD, respectively, while CMAP alone yields gains of $1.40$ and $0.50$ points. Their combination reaches $26.10$\% and $27.40$\% rare-category AP. Adding PTCD provides further gains of $0.60$ and $1.20$ points, bringing $\mathrm{AP}_{r}$ to $26.70$\% and $28.60$\%. The complete model performs best across rare, common, and frequent categories, with overall $\mathrm{mAP}$ reaching $36.50$\% and $37.70$\%, respectively.

Across both benchmarks, TGME yields larger novel-category gains than CMAP when added individually, while their combination outperforms either module alone. PTCD provides further improvements through semantically guided lateral correction and progressive multi-scale aggregation. These results support the effectiveness of the complete design across both detection-adapted frameworks.

\subsubsection{Effect of TGME}
We compare three strategies for incorporating textual semantics: Region-Text Matching (RTM), Cross-Attention Fusion (CAF), and TGME. RTM computes similarities between region and category text features, while CAF directly interacts visual and textual features through cross-attention. TGME instead combines query-specific text information, masked max-pooled semantics, and initial queries, then fuses the resulting queries with projected object tokens. These semantic queries guide subsequent cross-modal alignment in CMAP and multi-scale decoding in PTCD.
\begin{table*}[!thbp]
\caption{Comparison of text semantic modeling strategies on COCO-OVD and LVIS-OVD. RTM, CAF, and TGME denote Region--Text Matching, Cross-Attention Fusion, and the Text-Guided Modality Encoder, respectively. All results are reported in \%. $^{*}$ marks the detection-adapted OVSeg and SED baselines. Bold indicates the best result within each framework; red subscripts report gains in percentage points over the corresponding baseline.}\label{Tab3}
\centering
\renewcommand\arraystretch{1.0}
\resizebox{0.85\linewidth}{!}{
\normalsize
\begin{tabular}{c|ccc|ccc|cccc}
\hline
\hline
\multirow{2}{*}{Methods} & \multirow{2}{*}{RTM} & \multirow{2}{*}{CAF} & \multirow{2}{*}{TGME} & \multicolumn{3}{c|}{COCO-OVD(\%)} &  \multicolumn{4}{c}{LVIS-OVD(\%)} \\
& & & & AP$_{50}^\textrm{novel}$ & AP$_{50}^\textrm{base}$ &  AP$_{50}^\textrm{all}$ & AP$_r$ & AP$_c$ &  AP$_f$ & mAP\\
\hline
OVSeg-OVD$^{*}$  & \xmark & \xmark & \xmark & 33.70 & 53.70 & 49.30 & 22.40 & 28.30 & 36.00 & 32.40\\
\cdashline{1-11}
\multirow{3}{*}{TMCNet$_{OVSeg}$} & \cmark & \xmark & \xmark & 37.00$_{\color{red}\textrm{+3.30}}$	& 56.50	& 51.00	& 23.60$_{\color{red}\textrm{+1.20}}$	& 33.70 & 36.90 & 33.40\\
\multirow{3}{*}  & \xmark & \cmark & \xmark & 37.70$_{\color{red}\textrm{+4.00}}$ & 57.00 & 51.50	& 23.90$_{\color{red}\textrm{+1.50}}$ & 34.10 & 37.30 & 33.80\\
\multirow{3}{*}  & \xmark & \xmark & \cmark  & \textbf{39.20}$_{\color{red}\textrm{+5.50}}$	& \textbf{58.50} & \textbf{53.30} & \textbf{26.70}$_{\color{red}\textrm{+4.30}}$	& \textbf{37.10}	& \textbf{40.00}	& \textbf{36.50}\\

\hline
SED-OVD$^{*}$  & \cmark & \xmark & \xmark & 36.10 & 54.20 & 49.90 & 24.80 & 34.50 & 38.00 & 34.00\\
\cdashline{1-11}
\multirow{3}{*}{TMCNet$_{SED}$} & \cmark & \xmark & \xmark & 38.70$_{\color{red}\textrm{+2.60}}$ & 57.90 & 53.00 & 25.00$_{\color{red}\textrm{+0.20}}$ & 35.10 & 38.20 & 34.60\\
\multirow{3}{*} & \xmark & \cmark & \xmark & 39.80$_{\color{red}\textrm{+3.70}}$ & 59.20 & 53.10 & 25.50$_{\color{red}\textrm{+0.70}}$ & 35.30 & 38.70 & 34.80\\
\multirow{3}{*} & \xmark & \xmark & \cmark  & \textbf{41.60}$_{\color{red}\textrm{+5.50}}$ & \textbf{60.90} & \textbf{55.40} & 
\textbf{28.60}$_{\color{red}\textrm{+3.80}}$	& \textbf{38.10}	& \textbf{41.00} & \textbf{37.70}\\

\hline\hline
\end{tabular}
}
\end{table*}

\begin{table*}[!thbp]
\caption{Comparison of cross-modal alignment schemes on COCO-OVD and LVIS-OVD. CLIP, LP, and CMAP denote CLIP feature matching, linear projection, and the proposed Cross-Modal Alignment Pipeline, respectively. All results are reported in \%. $^{*}$ marks the detection-adapted OVSeg and SED baselines. Bold indicates the best result within each framework; red subscripts report gains in percentage points over the corresponding baseline.}\label{Tab4}
\centering
\renewcommand\arraystretch{1.0}
\resizebox{0.85\linewidth}{!}{
\normalsize
\begin{tabular}{c|ccc|ccc|cccc}
\hline
\hline
\multirow{2}{*}{Methods} & \multirow{2}{*}{CLIP} & \multirow{2}{*}{LP} & \multirow{2}{*}{CMAP} & \multicolumn{3}{c|}{COCO-OVD(\%)} &  \multicolumn{4}{c}{LVIS-OVD(\%)} \\
& & & & AP$_{50}^\textrm{novel}$ & AP$_{50}^\textrm{base}$ &  AP$_{50}^\textrm{all}$ & AP$_r$ & AP$_c$ &  AP$_f$ & mAP\\
\hline
OVSeg-OVD$^{*}$  & \cmark & \xmark & \xmark &33.70 & 53.70 & 49.30 & 22.40 & 28.30 & 36.00 & 32.40\\
\cdashline{1-11}
\multirow{3}{*}{TMCNet$_{OVSeg}$} & \cmark & \xmark & \xmark & 37.50$_{\color{red}\textrm{+3.80}}$	& 57.10	& 51.60	&24.00$_{\color{red}\textrm{+1.60}}$	& 34.10 & 37.60	&34.10\\
\multirow{3}{*}  & \xmark & \cmark & \xmark & 38.40$_{\color{red}\textrm{+4.70}}$	& 57.70 & 52.30	& 24.50$_{\color{red}\textrm{+2.10}}$	&34.60	& 38.00	& 34.50\\
\multirow{3}{*}  & \xmark & \xmark & \cmark  & \textbf{39.20}$_{\color{red}\textrm{+5.50}}$	& \textbf{58.50} & \textbf{53.30}	& 
\textbf{26.70}$_{\color{red}\textrm{+4.30}}$ & \textbf{37.10} & \textbf{40.00}	& \textbf{36.50}\\
\hline
SED-OVD$^{*}$  & - & \xmark & \xmark & 36.10 & 54.20 & 49.90 & 24.80 & 34.50 & 38.00 & 34.00\\
\cdashline{1-11}
\multirow{3}{*}{TMCNet$_{SED}$} 	& \cmark & \xmark & \xmark & 39.30$_{\color{red}\textrm{+3.20}}$ & 58.40 & 53.50 & 25.40$_{\color{red}\textrm{+0.60}}$ & 35.70 & 38.70 & 35.00\\
\multirow{3}{*} & \xmark & \cmark & \xmark & 40.20$_{\color{red}\textrm{+4.10}}$ & 59.40 & 53.70 & 25.70$_{\color{red}\textrm{+0.90}}$ & 35.70 & 39.10 & 35.00\\
\multirow{3}{*} & \xmark & \xmark & \cmark  & \textbf{41.60}$_{\color{red}\textrm{+5.50}}$ & \textbf{60.90} & \textbf{55.40} & 
\textbf{28.60}$_{\color{red}\textrm{+3.80}}$	& \textbf{38.10}	& \textbf{41.00} & \textbf{37.70}\\
\hline\hline
\end{tabular} }
\end{table*}

Table~\ref{Tab3} reports results for both detection-adapted frameworks, with their baselines included for reference. On COCO-OVD, TGME achieves $\mathrm{AP}_{50}^{\mathrm{novel}}$ scores of $39.20$\% and $41.60$\% for the OVSeg-OVD and SED-OVD models, respectively. These results exceed RTM by $2.20$ and $2.90$ percentage points and CAF by $1.50$ and $1.80$ points. TGME also achieves the highest base- and all-category AP in both frameworks.
On LVIS-OVD, TGME reaches $\mathrm{AP}_{r}$ scores of $26.70$\% and $28.60$\%, outperforming RTM by $3.10$ and $3.60$ points and CAF by $2.80$ and $3.10$ points, respectively. Improvements also extend to common and frequent categories, yielding the highest overall $\mathrm{mAP}$ of $36.50$\% and $37.70$\%.
These consistent gains support encoding textual priors into queries to guide subsequent visual feature processing. Compared with direct region-text matching and cross-attention fusion, the TGME configuration improves novel-category detection while maintaining stronger performance across the full evaluated vocabulary.

\subsubsection{Effect of CMAP}
\begin{figure}[pos=!tbp]
\centering
\includegraphics[width=.48\textwidth]{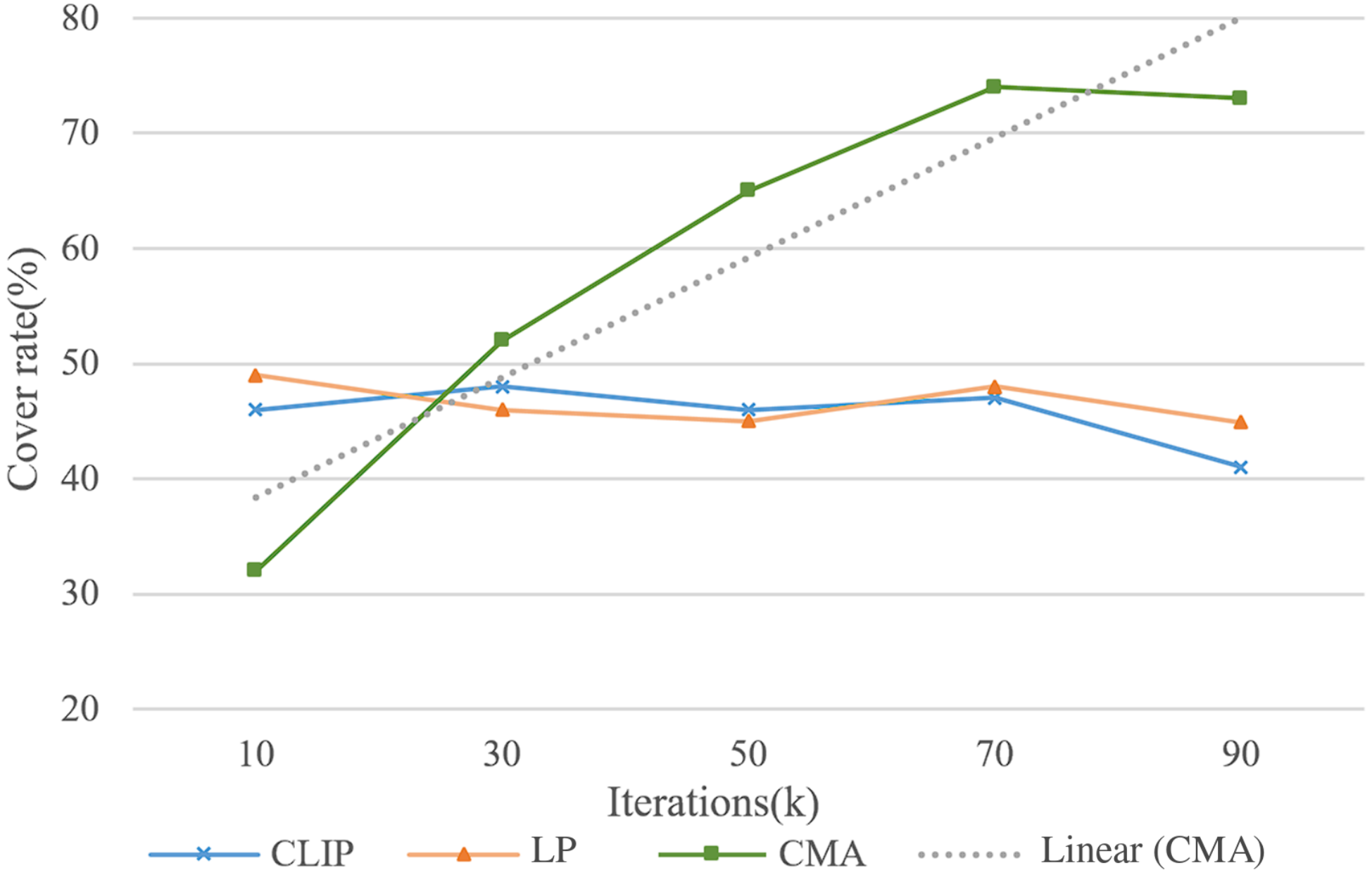}
\caption{Novel-category pseudo-label coverage on the COCO-OVD validation set across training iterations for different cross-modal alignment schemes. The horizontal axis shows training iterations in thousands, and the vertical axis shows coverage in \%.}
\label{fig3}
\vspace{-4mm}
\end{figure}

CMAP incorporates TGME queries into two complementary pathways: ALM supplies category semantics for lateral feature correction, while FO and UPM integrate the aligned queries with image-projected vision-text association features for progressive aggregation.
With TGME incorporated, Table~\ref{Tab4} compares CLIP feature matching, linear projection (LP), and CMAP. On COCO-OVD, CMAP achieves $\mathrm{AP}_{50}^{\mathrm{novel}}$ scores of $39.20$\% and $41.60$\% in the OVSeg-OVD and SED-OVD frameworks, respectively, exceeding CLIP by $1.70$ and $2.30$ percentage points and LP by $0.80$ and $1.40$ points. On LVIS-OVD, the corresponding $\mathrm{AP}_{r}$ scores reach $26.70$\% and $28.60$\%, surpassing CLIP by $2.70$ and $3.20$ points and LP by $2.20$ and $2.90$ points. CMAP also achieves the highest scores on the remaining metrics in both frameworks, supporting its effectiveness in incorporating textual guidance into visual feature processing.
Fig.~\ref{fig3} further compares novel-category pseudo-label coverage during training on COCO-OVD. CMAP increases coverage from approximately $32$\% at $10$\,k iterations to $74$\% at $70$\,k, remaining near $73$\% at $90$\,k. It exceeds both alternatives at every reported checkpoint from $30$\,k onward, while CLIP and LP remain within approximately $41$\%-$49$\%. This trend indicates broader coverage of novel-category objects and is consistent with the detection gains, although coverage alone does not measure pseudo-label precision.

\subsubsection{Effect of Detection Head}
We compare three detection configurations within the OVSeg-OVD and SED-OVD frameworks, with TGME, CMAP, and PTCD incorporated: CLIP feature matching (CLIP), Deformable DETR (Def\_DETR), and the proposed query-based CLIP\_Trans head. 
As shown in Table~\ref{Tab5}, CLIP\_Trans achieves $\mathrm{AP}_{50}^{\mathrm{novel}}$ scores of $39.20$\% and $41.60$\% on COCO-OVD in the OVSeg-OVD and SED-OVD frameworks, respectively. These scores exceed CLIP by $1.20$ and $2.20$ percentage points and Def\_DETR by $0.40$ and $0.90$ points. On LVIS-OVD, the corresponding $\mathrm{AP}_{r}$ scores reach $26.70$\% and $28.60$\%, surpassing CLIP by $2.70$ and $3.10$ points and Def\_DETR by $1.80$ and $2.60$ points.
CLIP\_Trans also achieves the highest scores on all remaining metrics in both frameworks. These results support its use for instance-level classification and localization with the text-guided features produced by TMCNet.
\begin{table*}[!thbp]
\caption{Comparison of detection configurations on COCO-OVD and LVIS-OVD, with TGME, CMAP, and PTCD incorporated in both frameworks. CLIP, Def\_DETR, and CLIP\_Trans denote the CLIP feature-matching configuration, the Deformable DETR-based configuration, and the adopted query-based CLIP--Transformer head, respectively. \cmark and \xmark indicate whether a configuration is used. All results are reported in \%. Bold indicates the best result within each framework; red subscripts report gains in percentage points over the corresponding OVSeg-OVD or SED-OVD baseline.}\label{Tab5}
\centering
\resizebox{0.85\linewidth}{!}{
\normalsize
\begin{tabular}{c|ccc|ccc|cccc}
\hline
\hline
\multirow{2}{*}{Detection-adapted Methods} & \multirow{2}{*}{CLIP} & \multirow{2}{*}{Def\_DETR} & \multirow{2}{*}{CLIP\_Trans} & \multicolumn{3}{c|}{COCO-OVD(\%)} & \multicolumn{4}{c}{LVIS-OVD(\%)} \\
& & & & AP$_{50}^\textrm{novel}$ & AP$_{50}^\textrm{base}$ & AP$_{50}^\textrm{all}$ & AP$_r$ & AP$_c$ & AP$_f$ & mAP\\
\hline
\multirow{3}{*}{TMCNet$_{OVSeg}$} & \cmark & \xmark & \xmark & 38.00 & 57.30 & 52.00 & 24.00 & 34.00 & 37.30 & 33.50\\
& \xmark & \cmark & \xmark & 38.80 & 58.00 & 52.80 & 24.90 & 35.00 & 38.20 & 34.60\\
& \xmark & \xmark & \cmark & \textbf{39.20}$_{\color{red}\textrm{+1.20}}$ & \textbf{58.50} & \textbf{53.30} &
\textbf{26.70}$_{\color{red}\textrm{+4.30}}$	& \textbf{37.10}	& \textbf{40.00}	& \textbf{36.50}\\
\cdashline{1-11}
\multirow{3}{*}{TMCNet$_{SED}$} & \cmark & \xmark & \xmark & 39.40 & 58.50 & 52.90 & 25.50 & 35.20 & 38.80 & 34.80\\
& \xmark & \cmark & \xmark & 40.70 & 59.90 & 54.30 & 26.00 & 35.90 & 39.40 & 35.50\\
& \xmark & \xmark & \cmark & \textbf{41.60}$_{\color{red}\textrm{+2.20}}$ & \textbf{60.90} & \textbf{55.40} & 
\textbf{28.60}$_{\color{red}\textrm{+3.80}}$	& \textbf{38.10}	& \textbf{41.00} & \textbf{37.70}\\
\hline\hline
\end{tabular}
}
\end{table*}

\subsection{Comparisons with State-of-the-Art Methods}\label{sec4:4}
Tables~\ref{Tab6} and~\ref{Tab7} compare TMCNet with representative open-vocabulary detectors on COCO-OVD and LVIS-OVD, respectively. Fig.~\ref{fig4} provides qualitative comparisons.

\subsubsection{Detection Accuracy and Inference Efficiency}
\textbf{COCO-OVD.}
Table~\ref{Tab6} shows that TMCNet$_{\mathrm{OVSeg}}$ and TMCNet$_{\mathrm{SED}}$ achieve $\mathrm{AP}_{50}^{\mathrm{novel}}$ scores of $39.20$\% and $41.60$\%, respectively, both improving their baselines by $5.50$ percentage points. TMCNet$_{\mathrm{SED}}$ leads the listed V-OVD methods across all three metrics, exceeding CAKE~\cite{ma2025cake} by $3.40$ points on novel categories.
These results show improved novel-category detection alongside gains on base categories across both frameworks.

\textbf{LVIS-OVD.}
In Table~\ref{Tab7}, TMCNet$_{\mathrm{OVSeg}}$ and TMCNet$_{\mathrm{SED}}$ achieve $\mathrm{AP}_{r}$ scores of $26.70$\% and $28.60$\%, exceeding their baselines by $4.30$ and $3.80$ points, respectively. TMCNet$_{\mathrm{SED}}$ achieves the highest common-category AP, frequent-category AP, and overall mAP among the listed methods, with mAP reaching $37.70$\%, although BIND retains the highest rare-category AP. Under the reported configurations, the two TMCNet implementations reduce baseline latency by $11.38$\% and $63.59$\%, respectively.
These results demonstrate gains across category frequency groups and lower inference latency relative to both baselines, with particularly strong overall accuracy for TMCNet$_{\mathrm{SED}}$.
\begin{table*}[!thbp]
\caption{Comparison of TMCNet with representative open-vocabulary object detection methods on COCO-OVD. $^{\dagger}$ denotes results obtained by retraining and evaluating the models in this work. $^{*}$ indicates the detection-adapted variants of OVSeg and SED. Def\_DETR denotes the Deformable DETR-based detection configuration, and CLIP\_Trans denotes the query-based CLIP+Transformer detection head adopted in this work. $-$ denotes unreported results}\label{Tab6}
\centering
\renewcommand\arraystretch{1.0}
\resizebox{0.85\linewidth}{!}{
\normalsize
\begin{tabular}{c|c|c|c|c|ccc}
\hline\hline
\multirow{2}{*}{Methods} & \multirow{2}{*}{Publication} & \multirow{2}{*}{Train Setting} & \multirow{2}{*}{Backbone} & \multirow{2}{*}{detectors} & \multicolumn{3}{c}{COCO-OVD(\%)} \\
& & & & & AP$_{50}^{\mathrm{novel}}$ & AP$_{50}^{\mathrm{base}}$ & AP$_{50}^{\mathrm{all}}$ \\
\hline
OVR-CNN~\cite{zareian2021open}$^{\dagger}$ & CVPR'21 & V-OVD & RN50-C4 & Faster R-CNN & 22.80 & 46.00 & 39.90 \\
BARON~\cite{wu2023aligning}$^{\dagger}$ & CVPR'23 & V-OVD & RN50-C4 & Faster R-CNN & 33.10 & 54.80 & 49.10 \\
BIND~\cite{zhang2024exploring}$^{\dagger}$ & CVPR'24 & V-OVD & ViT-B/16 & Built-in detector & 36.30 & 54.70 & 50.20 \\
CAKE~\cite{ma2025cake}$^{\dagger}$ & AAAI'25 & V-OVD & RN50 & Faster R-CNN & 38.20 & 58.00 & 52.80 \\
\cdashline{1-8}
RegionCLIP~\cite{zhong2022regionclip}$^{\dagger}$ & CVPR'22 & C-OVD & RN50-C4 & CLIP & 26.80 & 54.80 & 47.50 \\
CoDet~\cite{ma2023codet}$^{\dagger}$ & NeurIPS'23 & C-OVD & RN50 & CLIP & 30.60 & 52.30 & 46.60 \\
CAKE~\cite{ma2025cake}$^{\dagger}$ & AAAI'25 & C-OVD & RN50 & Faster R-CNN & 41.30 & 60.20 & 55.30 \\
\cdashline{1-8}
OV-DETR~\cite{zang2022open}$^{\dagger}$ & ECCV'22 & G-OVD & RN50 & Def\_DETR & 29.40 & 61.00 & 52.70 \\
CORA~\cite{wu2023cora}$^{\dagger}$ & CVPR'23 & G-OVD & RN50 & Def\_DETR & 35.10 & 35.50 & 35.40 \\
CAKE~\cite{ma2025cake}$^{\dagger}$ & AAAI'25 & G-OVD & RN50 & Faster R-CNN & 39.10 & 58.10 & 53.10 \\
\cdashline{1-8}
Detic~\cite{zhou2022detecting}$^{\dagger}$ & ECCV'22 & WS-OVD & Swin-B & Faster R-CNN & 28.40 & 53.80 & 47.20 \\
GOAT~\cite{wang2023open}$^{\dagger}$ & ICCV'23 & WS-OVD & RN50 & Faster R-CNN & 36.40 & 53.00 & 48.60 \\
SAS-Det~\cite{zhao2024taming}$^{\dagger}$ & CVPR'24 & WS-OVD & RN50-C4 & Faster R-CNN & 37.40 & 58.50 & 53.00 \\
CAKE~\cite{ma2025cake}$^{\dagger}$ & AAAI'25 & WS-OVD & RN50 & Faster R-CNN & 41.80 & 60.60 & 55.70 \\
\cdashline{1-8}
MSPL~\cite{choi2026mspl}$^{\dagger}$ & ECCV'26 & Pseudo-labels & RN50-FPN & Faster R-CNN & 39.00 & 55.30 & -- \\
BARON-BFDet~\cite{li2026bfdet}$^{\dagger}$ & AAAI'26 & Pseudo-labels & RN50-C4 & Faster R-CNN
& 37.80 & 58.20 & 52.00 \\
ViLD-BFDet~\cite{li2026bfdet}$^{\dagger}$ & AAAI'26 & Pseudo-labels & RN50-C4 & Faster R-CNN
& 29.90 & 62.10 & 53.70 \\
RegionCLIP-BFDet~\cite{li2026bfdet}$^{\dagger}$ & AAAI'26 & Pseudo-labels & RN50-C4 & Faster R-CNN
& 29.50 & 58.00 & 50.40 \\
VLDet-BFDet~\cite{li2026bfdet}$^{\dagger}$ & AAAI'26 & Pseudo-labels & RN50-C4 & Faster R-CNN
& 35.10 & 54.10 & 49.00 \\
\cdashline{1-8}
OVSeg-OVD$^{*}$ & CVPR'23 & V-OVD & ViT-B/16 & CLIP\_Trans & 33.70 & 53.70 & 49.30 \\
SED-OVD$^{*}$ & CVPR'24 & V-OVD & ConvNeXt-B & CLIP\_Trans & 36.10 & 54.20 & 49.90 \\
\cdashline{1-8}
TMCNet$_{\mathrm{OVSeg}}$ & None & V-OVD & ViT-B/16 & CLIP\_Trans & \textbf{39.20} & \textbf{58.50} & \textbf{53.30} \\
TMCNet$_{\mathrm{SED}}$ & None & V-OVD & ConvNeXt-B & CLIP\_Trans & \textbf{41.60} & \textbf{60.90} & \textbf{55.40} \\
\hline\hline
\end{tabular}
}
\end{table*}

\begin{table*}[!thbp]
\caption{Comparison of open-vocabulary object detection methods on LVIS-OVD. $\dagger$ denotes models retrained by the authors of this work. $^{*}$ indicates the detection-adapted variants of OVSeg and SED. Def\_DETR denotes the Deformable DETR-based detection configuration, and CLIP\_Trans denotes the query-based CLIP+Transformer detection head adopted in this work. $-$ denotes an unreported result.}\label{Tab7}
\centering
\renewcommand\arraystretch{1.0}
\resizebox{0.85\linewidth}{!}{
\normalsize
\begin{tabular}{c|c|c|c|cccc|c}
\hline\hline
\multirow{2}{*}{Methods} & \multirow{2}{*}{Publication} & \multirow{2}{*}{Backbone} & \multirow{2}{*}{detectors} & \multicolumn{4}{c|}{LVIS-OVD(\%)} & \multirow{2}{*}{Inf. time (ms/image)} \\
& & & & AP$_r$ & AP$_c$ & AP$_f$ & mAP & \\
\hline
OV-DETR~\cite{zang2022open}$^{\dagger}$ & ECCV'22 & RN50 & Def\_DETR & 21.00 & 25.00 & 32.50 & 26.60 & 574.20 \\
BIND~\cite{zhang2024exploring}$^{\dagger}$ & CVPR'24 & ViT-B/16 & Built-in detector & 29.40 & 30.60 & 33.50 & 31.40 & 42.60 \\
SAS-Det~\cite{zhao2024taming}$^{\dagger}$ & CVPR'24 & RN50-C4 & Faster R-CNN & 20.90 & 26.10 & 31.60 & 27.40 & 71.20 \\
CAKE~\cite{ma2025cake}$^{\dagger}$ & AAAI'25 & RN50 & Faster R-CNN & 25.00 & 34.80 & 38.40 & 34.90 & 35.80 \\

\cdashline{1-9}
MSPL~\cite{choi2026mspl}$^{\dagger}$ & ECCV'26 & RN50-FPN & Faster R-CNN & 25.00 & 33.30 & 37.20 & 33.50 & - \\
GenRep-LDM~\cite{li2026genrep}$^{\dagger}$ & ICLR'26 & LDM & - & 27.90 & 30.60 & 33.10 & 32.00 & - \\
GenRep-DiT~\cite{li2026genrep}$^{\dagger}$ & ICLR'26 & DiT & - & 28.00 & 30.10 & 35.80 & 34.60 & - \\

\cdashline{1-9}
OVSeg-OVD$^{*}$ & CVPR'23 & ViT-B/16 & CLIP\_Trans & 22.40 & 28.30 & 36.00 & 32.40 & 240.00 \\
SED-OVD$^{*}$ & CVPR'24 & ConvNeXt-B & CLIP\_Trans & 24.80 & 34.50 & 38.00 & 34.00 & 64.00 \\
\cdashline{1-9}
TMCNet$_{\mathrm{OVSeg}}$ & None & ViT-B/16 & CLIP\_Trans & \textbf{26.70}	& \textbf{37.10}	& \textbf{40.00}	& \textbf{36.50} & \textbf{212.70} \\
TMCNet$_{\mathrm{SED}}$ & None & ConvNeXt-B & CLIP\_Trans & \textbf{28.60}	& \textbf{38.10}	& \textbf{41.00} & \textbf{37.70} & \textbf{23.30} \\
\hline\hline
\end{tabular}
}
\end{table*}

\subsubsection{Qualitative Results on COCO-OVD}
\begin{figure*}[pos=b]
\centering
\includegraphics[width=.90\textwidth]{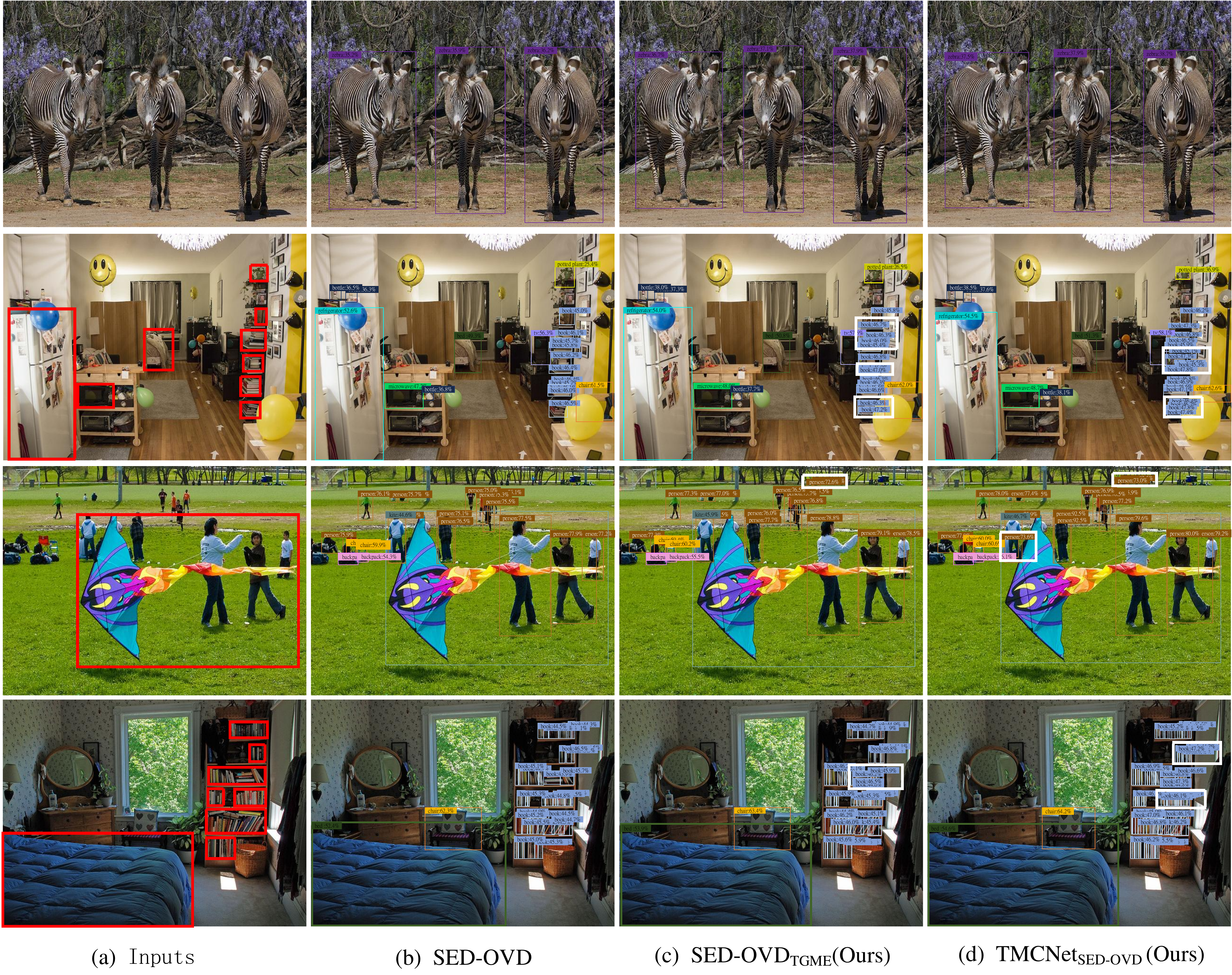}
\caption{Qualitative detection results for SED-OVD and its variants on COCO-OVD. From left to right: input images and predictions from SED-OVD, SED-OVD$_{\mathrm{TGME}}$, and TMCNet$_{\mathrm{SED-OVD}}$. Red solid boxes in the input images highlight selected novel-category objects, while white solid boxes in the predictions highlight additional objects detected relative to SED-OVD. Colored bounding boxes and their labels indicate predicted locations, categories, and confidence scores.}
\label{fig4}
\end{figure*} 

Fig.~\ref{fig4} presents four examples of detection results from SED-OVD, its TGME-augmented variant SED-OVD$_{\mathrm{TGME}}$, and the full model TMCNet$_{\mathrm{SED-OVD}}$ on COCO-OVD. These examples include scenes with multiple objects, occlusion, and densely arranged objects.
In the first example, all three models detect the main zebra instances, while incorporating TGME and the full set of modules increases the prediction confidence for some instances. In the second example, an indoor scene, the proposed variants recover several book instances missed by SED-OVD. They also detect objects from novel categories, including refrigerators, microwaves, beds, and potted plants, and accurately localize occluded bottles. In the third example, all models detect the kite, while the proposed variants identify additional distant or partially occluded people. In the fourth example, the proposed variants detect more of the densely arranged books on the shelves.
These visual comparisons illustrate fewer missed detections in several challenging scenes. The observations are consistent with the quantitative results and provide complementary evidence for the effectiveness of the proposed text-guided modality correction mechanism.

\subsection{Extension to Open-Vocabulary Semantic Segmentation}\label{sec4:5}
To assess its applicability to dense prediction, we integrate TMCNet into OVSeg~\cite{liang2023open} and SED~\cite{xie2024sed} while retaining their segmentation prediction mechanisms. Both models are trained on COCO-Stuff-$171$ with pixel-level annotations and evaluated using mIoU under five settings across ADE20K, PASCAL-Context, and PASCAL VOC.

\begin{table*}[b]
\caption{Effectiveness of TMCNet in different open-vocabulary semantic segmentation frameworks. $^{\dagger}$ denotes results obtained by retraining and evaluating the models in this work. $-$ indicates that no additional feature backbone is used. Boldface highlights our methods, and red subscripts indicate absolute gains in percentage points over the corresponding reproduced baseline.}\label{Tab8}
\centering
\renewcommand\arraystretch{1.0}
\resizebox{0.85\linewidth}{!}{
\normalsize
\begin{tabular}{c|c|c|c|ccccc}
\hline
\hline
Methods & VLM & Feature backbone & Train Dataset & A-847 & PC-459 & A-150 &  PC-59 & PAS-20\\
\hline
OVSeg~\cite{liang2023open}$^{\dagger}$ & ViT-B/16 & Swin-B & COCO-Stuff-171 & 7.10 & 11.00 & 24.80 & 53.30 & 92.60 \\
TMCNet$_{OVSeg}$ & ViT-B/16 & Swin-B & COCO-Stuff-171 & \textbf{9.10}$_{\color{red}\textrm{+2.00}}$ & \textbf{14.20}$_{\color{red}\textrm{+3.20}}$	& \textbf{30.90}$_{\color{red}\textrm{+6.10}}$ &	\textbf{55.70}$_{\color{red}\textrm{+2.40}}$ & \textbf{95.40}$_{\color{red}\textrm{+2.80}}$ \\
\cdashline{1-9}
SED~\cite{xie2024sed}$^{\dagger}$ & ConvNeXt-B & - & COCO-Stuff-171 & 8.90 & 12.40 & 31.20 & 47.30 & 93.70 \\
TMCNet$_{SED}$ & ConvNeXt-B & - & COCO-Stuff-171 & \textbf{11.40}$_{\color{red}\textrm{+2.50}}$ & \textbf{16.70}$_{\color{red}\textrm{+4.30}}$ & \textbf{32.80}$_{\color{red}\textrm{+1.60}}$ & \textbf{57.40}$_{\color{red}\textrm{+10.10}}$ & \textbf{95.30}$_{\color{red}\textrm{+1.60}}$\\
\hline\hline
\end{tabular}
}
\end{table*}

\textbf{Effectiveness of TMCNet.}
As shown in Table~\ref{Tab8}, TMCNet improves both reproduced baselines across all five settings, with gains of $2.00$-$6.10$ percentage points for OVSeg and $1.60$-$10.10$ points for SED. The largest gains occur on A-$150$ and PC-$59$, respectively. Improvements on A-$847$ and PC-$459$ also demonstrate benefits for pixel-level prediction with large category vocabularies.
\begin{table*}[!thbp]
\caption{Comparison with representative open-vocabulary semantic segmentation methods. All results are reported in mIoU (\%). $\dagger$ denotes results reproduced in this work; other baseline results are taken from the corresponding papers or official implementations. $-$ denotes an unreported result.}
\label{Tab9}
\centering
\renewcommand\arraystretch{1.0}
\resizebox{0.85\linewidth}{!}{
\normalsize
\begin{tabular}{c|c|c|c|ccccc}
\hline
\hline
Methods & Publication & VLM & Training Dataset
& A-847 & PC-459 & A-150 & PC-59 & PAS-20 \\
\hline
ZegFormer~\cite{ding2022decoupling}
& CVPR'22 & ViT-B/16 & COCO-Stuff-156
& 4.90 & 9.10 & 16.90 & 42.80 & 86.20 \\
ZSseg~\cite{xu2022simple}
& ECCV'22 & ViT-B/16 & COCO-Stuff-171
& 7.00 & -- & 20.50 & 47.70 & 88.40 \\
OVSeg~\cite{liang2023open}$\dagger$
& CVPR'23 & ViT-B/16 & COCO-Stuff-171
& 7.10 & 11.00 & 24.80 & 53.30 & 92.60 \\
SED~\cite{xie2024sed}$^{\dagger}$
& CVPR'24 & ConvNeXt-B & COCO-Stuff-171
& 8.90 & 12.40 & 31.20 & 47.30 & 93.70 \\
CAT-Seg + DeCLIP~\cite{wang2025declip}$^{\dagger}$
& CVPR'25 & ViT-B/16 & COCO-Stuff-171
& 9.50 & 15.40 & 31.30 & 55.60 & 92.60 \\
LoGoSeg~\cite{chen2026logoseg}$^{\dagger}$
& AAAI'26 & ConvNeXt-B & COCO-Stuff-171
& 10.20 & 16.30 & 32.00 & 55.80 & 93.00 \\
LoGoSeg~\cite{chen2026logoseg}$^{\dagger}$
& AAAI'26 & ViT-B/16 & COCO-Stuff-171
& 10.60 & 16.50 & 32.60 & 56.20 & 93.40 \\
\cdashline{1-9}
TMCNet$_{\mathrm{OVSeg}}$
& Ours & ViT-B/16 & COCO-Stuff-171
& \textbf{9.10}
& \textbf{14.20}
& \textbf{30.90}
& \textbf{55.70}
& \textbf{95.40}\\
TMCNet$_{\mathrm{SED}}$
& Ours & ConvNeXt-B & COCO-Stuff-171
& \textbf{11.40}
& \textbf{16.70}
& \textbf{32.80}
& \textbf{57.40}
& \textbf{95.30} \\
\hline
\hline
\end{tabular}
}
\end{table*}

\textbf{Comparison with Representative Methods.}
Table~\ref{Tab9} shows that TMCNet$_{\mathrm{SED}}$ achieves the highest mIoU among the listed configurations on four settings, while TMCNet$_{\mathrm{OVSeg}}$ leads on PAS-$20$ with $95.40$\%. Across all five settings, TMCNet$_{\mathrm{SED}}$ outperforms the reproduced CAT-Seg + DeCLIP~\cite{wang2025declip} by $1.30$-$2.70$ points and LoGoSeg~\cite{chen2026logoseg} with the same ConvNeXt-B visual encoder by $0.40$-$2.30$ points. These results support the applicability of text-guided modality correction to open-vocabulary semantic segmentation across both frameworks.

\section{Conclusions}
In this work, we present TMCNet, a Text-Guided Modality Correction Network for open-vocabulary object detection that incorporates category semantics into query construction, cross-modal alignment, and multiscale decoding. The Text-Guided Modality Encoder uses multi-head cross-attention and masked max pooling to encode textual priors into initial queries, which are combined with projected object tokens to form semantic query representations. The Cross-Modal Alignment Pipeline maps these queries into a shared feature space and provides two complementary outputs: dimension-adapted queries for lateral visual feature correction, and joint features obtained by fusing aligned queries with projected vision–text association features. The Progressive Text Correction Decoder then integrates the corrected lateral features and scale-adapted joint features with the evolving decoder representation, combining high-level semantics with spatial details for object classification and bounding box regression. Experiments on COCO-OVD and LVIS-OVD show improvements in novel-category and overall detection performance across both detection-adapted frameworks. Incorporating the same mechanism into OVSeg and SED also improves segmentation accuracy across all five evaluation settings, supporting its applicability to open-vocabulary pixel-level prediction.

\bibliographystyle{elsarticle-num}
\bibliography{cas-refs}
\end{document}